\documentclass{article}

\usepackage{graphicx} 
\usepackage{url}
\usepackage{multirow}
\usepackage{authblk}
\usepackage{float}
\usepackage{hyperref}
\usepackage{geometry}
\usepackage[small,compact]{titlesec}
\usepackage{tabularray}
\usepackage{listings}
\usepackage[sorting=none,maxnames=4,minnames=3, bibstyle=numeric, hyperref=true,doi=false,isbn=false,
firstinits=true,abbreviate=false,backend=bibtex8]{biblatex}

\bibliography{References}
\graphicspath{{figures/}}
\title{Seg-metrics: a Python package to compute segmentation metrics}

\author[1]{Jingnan Jia}
\author[1,*]{Marius Staring}
\author[1,*]{Berend C. Stoel}
\date{} 

\affil[1]{Division of Image Processing, Department of Radiology, Leiden University Medical Center, 2300RC, the Netherlands}

\begin{document}

\maketitle
\begin{abstract}

In response to a concerning trend of selectively emphasizing metrics in medical image segmentation (MIS) studies, we introduce \texttt{seg-metrics}, an open-source Python package for standardized MIS model evaluation. Unlike existing packages, \texttt{seg-metrics} offers user-friendly interfaces for various overlap-based and distance-based metrics, providing a comprehensive solution. \texttt{seg-metrics} supports multiple file formats and is easily installable through the Python Package Index (PyPI). With a focus on speed and convenience, \texttt{seg-metrics} stands as a valuable tool for efficient MIS model assessment.
\\
\end{abstract}

\section{Background}

In the last decade, the research of artificial intelligence on medical images has attracted researchers' interest. One of the most popular directions is automated medical image segmentation (MIS) using deep learning, which aims to automatically assign labels to pixels so that the pixels with the same label from a segmented object. However, in the past years a strong trend of highlighting or cherry-picking improper metrics to show particularly high scores close to 100\% was revealed in scientific publishing of MIS studies \cite{muller2022towards}. In addition, even though there are some papers that evaluate image segmentation results from different perspectives, the implementation of their evaluation algorithms is inconsistent. This is due to the lack of a universal metric library in Python for standardized and reproducible evaluation. Therefore, we propose to develop an open-source publicly available Python package \texttt{seg-metrics}, which aims to evaluate the performance of MIS models.

\section{Related packages}

As far as we know, till the publish date of this package (2020), there are two open source packages which could perform MIS metrics calculation: \texttt{SimpleITK}\cite{lowekamp2013design} and \texttt{Medpy} \cite{Maier}.

\textbf{SimpleITK} is an interface (including Python, c\#, Java, and R) to the Insight Segmentation and Registration Toolkit (ITK) designed for biomedical image analysis. \texttt{SimpleITK} does not support the evaluation of MIS directly.  Each evaluation consists of several steps, which makes it not user friendly for users. \textbf{Medpy} is a medical image processing library  written in Python. It includes some functions to evaluate MIS. However, it mainly support the operations of binary segmentation results, which limits its wider application cenarios. Therefore, this work aims to develop a Python package specifically for MIS.

\section{Our \texttt{seg-metrics} package}

Our \texttt{seg-metrics} package supports to calculate different evaluation metrics directly in one line. The metrics could be divided to overlap-based metrics and distance-based metrics. Overlap-based metrics, measure the overlap between the reference annotation and the prediction of the algorithm. It is typically complemented by a distance-based metrics, which could explicitly assess how close the boundaries are between the prediction and the reference \cite{maier2022metrics}. The details of the two categories are described below.

\subsection{Overlap-based metrics}

A confusion matrix (see Table \ref{confusion_matrix}) could be derived when comparing a segmentation result and its reference. In this table, there are 4 different outcomes:

\begin{enumerate}
    \item \textbf{TP}: If the actual classification is positive and the predicted classification is positive, this is called a true positive result because the positive sample was correctly identified by the classifier. 
    
    \item \textbf{FN}: If the actual classification is positive and the predicted classification is negative, this is called a false negative result because the positive sample is incorrectly identified by the classifier as being negative. 
    
    \item \textbf{FP}: If the actual classification is negative and the predicted classification is positive, this is called a false positive result because the negative sample is incorrectly identified by the classifier as being positive. 
    
    \item \textbf{TN}: If the actual classification is negative and the predicted classification is negative, this is called a true negative result because the negative sample gets correctly identified by the classifier.
\end{enumerate}

\begin{table}  
\centering
\caption{Confusion matrix (adopted from \url{https://en.wikipedia.org/wiki/Confusion_matrix})}
\begin{tblr}{
  cells = {c},
  cell{1}{1} = {c=2,r=2}{},
  cell{1}{3} = {c=2}{},
  cell{3}{1} = {r=2}{},
  vlines,
  hline{1,3,5} = {-}{},
  hline{2} = {3-4}{},
  hline{4} = {2-4}{},
}
Total (P+N) &              & Prediction   &              \\
            &              & Positive (P) & Negative (N) \\
Reference   & Positive (P) & TP           & FN           \\
            & Negative (N) & FP           & TN           
\end{tblr}
\label{confusion_matrix}
\end{table}

Based on these four outcomes, we can derive a great number of overlap-based metrics. Their equations are as follows.

\begin{itemize}

\item Dice Coefficient (F1-Score)
\begin{equation}
Dice = \frac{2 \times |A \cap B|}{|A| + |B|} = \frac{2 \times TP}{2 \times TP + FP + FN}
\end{equation}

\item Jaccard index
\begin{equation}
Jaccard = \frac{|A \cap B|}{|A \cup B|} = \frac{TP}{TP + FP + FN}
\end{equation}

\item Precision/Positive predictive value (PPV)

Precision score is the number of true positive results divided by the number of all positive results
\begin{equation}
Precision = \frac{TP}{TP + FP}
\end{equation}

\item Selectivity/Specificity/True negative rate
\begin{equation}
Specificity = \frac{TN}{TN + FP}
\end{equation}

\item Recall/Sensitivity/Hit rate/True positive rate (TPR)

Recall score, also known as Sensitivity, hit rate, or TPR, is the number of true positive results divided by the number of all samples that should have been identified as positive

\begin{equation}
Sensitivity = \frac{TP}{TP + FN}
\end{equation}

\item Accuracy/Rand Index

Accuracy score, also known as Rand index is the number of correct predictions, consisting of correct positive and negative predictions divided by the total number of predictions.
\begin{equation}
    Accuracy = \frac{TP + TN}{TP + FP + FN + TN}
\end{equation}

\end{itemize}

\subsection{Distance-based metrics}
\begin{itemize}

\item Hausdorff distance (HD) (see Figure \ref{fig:hausdorff})
\begin{equation}
HD = \max \left\{ \sup_{a \in A} \inf_{b \in B} d(a, b), \sup_{b \in B} \inf_{a \in A} d(b, a) \right\}
\end{equation}
where $sup$ represents the supremum operator, $inf$ is the infimum operator, and $inf_{b \in B} d(a, b)$ quantifies the distance from a point $a \in X$ to the subset $B \subseteq X$.

\begin{figure}[tb]
    \centering
    \includegraphics[width=0.6\textwidth]{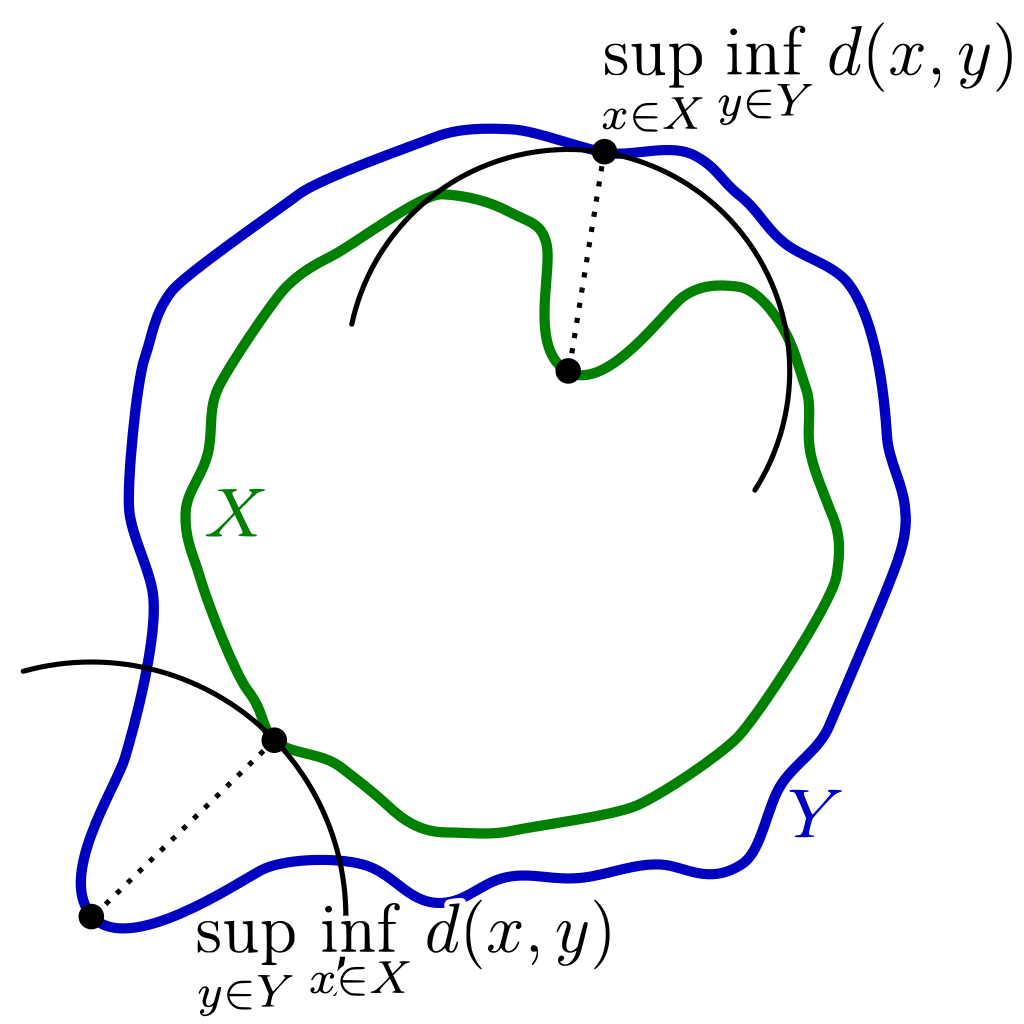}
    \caption{Hausdorff distance between the green curve X and the blue curve Y.}
    \label{fig:hausdorff}
\end{figure}

\item Hausdorff distance 95\% percentile (HD95)

\item Mean (Average) surface distance (MSD)

\item Median surface distance (MDSD)

\end{itemize}

\section{Installation}

Our package was published in the Python Package Index (PyPI), which is the official third-party software repository for Python. Thus, \texttt{seg-metrics} can be directly installed and immediately used in any Python environment using a single line as follows.  

\begin{lstlisting}[frame=single]
pip install seg-metrics
\end{lstlisting}

\section{Use cases}

\texttt{seg-metrics} is a Python package which output the segmentation metrics by receiving one ground truth image and another predicted image. After we import the package by \texttt{from seg\_metrics import seg\_metrics}, the syntax to use it is as follow (\textbf{Note:} all the following cases are based on texttt{seg-metrics 1.1.6}).

\begin{lstlisting}[frame=single]
write_metrics(labels,
              gdth_path = None,
              pred_path = None,
              csv_file = None,
              gdth_img = None,
              pred_img = None,
              metrics = None,
              verbose = False,
              spacing = None,
              fully_connected = True,
              TPTNFPFN = False)
    """ Parameter description.
    labels: a list of labels to performe the calculation of metrics.
    gdth_path: a (sequence of) path of ground truth.
    pred_path: a (sequence of) path of prediction.
    csv_file: filename to save the metrics.
    gdth_img: a (sequence of) ground truth.
    pred_img: a (sequence of) prediction.
    metrics: metric names.
    verbose: whether to show the animated progress bar
    spacing: spacing of input images.
    fully_connected: whether to apply fully connected border.
    TPTNFPFN: whether to return the confusion matrix.
    
    return: A dict or a list of dicts which store metrics.
    """
\end{lstlisting}

More examples are shown below.

\begin{itemize}
    \item Evaluate two batches of images with same filenames from two different folders.

\begin{lstlisting}[frame=single]

labels = [4, 5 ,6 ,7 , 8]
gdth_path = 'data/gdth'  # folder for ground truth images
pred_path = 'data/pred'  # folder for predicted images
csv_file = 'metrics.csv'  # file to save results

metrics = sg.write_metrics(labels=labels,
                  gdth_path=gdth_path,
                  pred_path=pred_path,
                  csv_file=csv_file)
print(metrics)  
\end{lstlisting}

        \item Evaluate two images

\begin{lstlisting}[frame=single]
labels = [4, 5 ,6 ,7 , 8]
gdth_file = 'data/gdth.mhd'  # full path for ground truth image 
pred_file = 'data/pred.mhd'  # full path for prediction image
csv_file = 'metrics.csv'

metrics = sg.write_metrics(labels=labels,
                           gdth_path=gdth_file,
                           pred_path=pred_file,
                           csv_file=csv_file)
\end{lstlisting}

        \item Evaluate two images with specific metrics

\begin{lstlisting}[frame=single]
labels = [0, 4, 5 ,6 ,7 , 8]
gdth_file = 'data/gdth.mhd'
pred_file = 'data/pred.mhd'
csv_file = 'metrics.csv'

metrics = sg.write_metrics(labels=labels[1:],
                  gdth_path=gdth_file,
                  pred_path=pred_file,
                  csv_file=csv_file,
                  metrics=['dice', 'hd'])
# for only one metric
metrics = sg.write_metrics(labels=labels[1:], 
                  gdth_path=gdth_file,
                  pred_path=pred_file,
                  csv_file=csv_file,
                  metrics='msd')  
\end{lstlisting}

\item Select specific metrics.
By passing the following parameters to select specific metrics.

\begin{lstlisting}[frame=single]
# ----------Overlap based metrics---------------
- dice:         Dice (F-1)
- jaccard:      Jaccard
- precision:    Precision
- recall:       Recall
- fpr:          False positive rate
- fnr:          False negtive rate
- vs:           Volume similarity
# ----------Distance based metrics---------------
- hd:           Hausdorff distance
- hd95:         Hausdorff distance 95% percentile
- msd:          Mean (Average) surface distance
- mdsd:         Median surface distance
- stdsd:        Std surface distance
\end{lstlisting}

For example:

\begin{lstlisting}[frame=single]
labels = [1]
gdth_file = 'data/gdth.mhd'
pred_file = 'data/pred.mhd'
csv_file = 'metrics.csv'

metrics = sg.write_metrics(labels, gdth_file, pred_file, 
                           csv_file, metrics=['dice', 'hd95'])
dice = metrics['dice']
hd95 = metrics['hd95']
\end{lstlisting}

\end{itemize}

\section{Comparison to other packages}
\texttt{medpy} also provide functions to calculate metrics for medical images. Compared to it, our package \texttt{seg-metrics} has several advantages.

\begin{itemize}
    \item \textbf{Faster.} \texttt{seg-metrics} is 5-10 times faster calculating distance based metrics (see Figure \ref{fig:comp}).

    \begin{figure}
        \centering
        \includegraphics[width=0.75\linewidth]{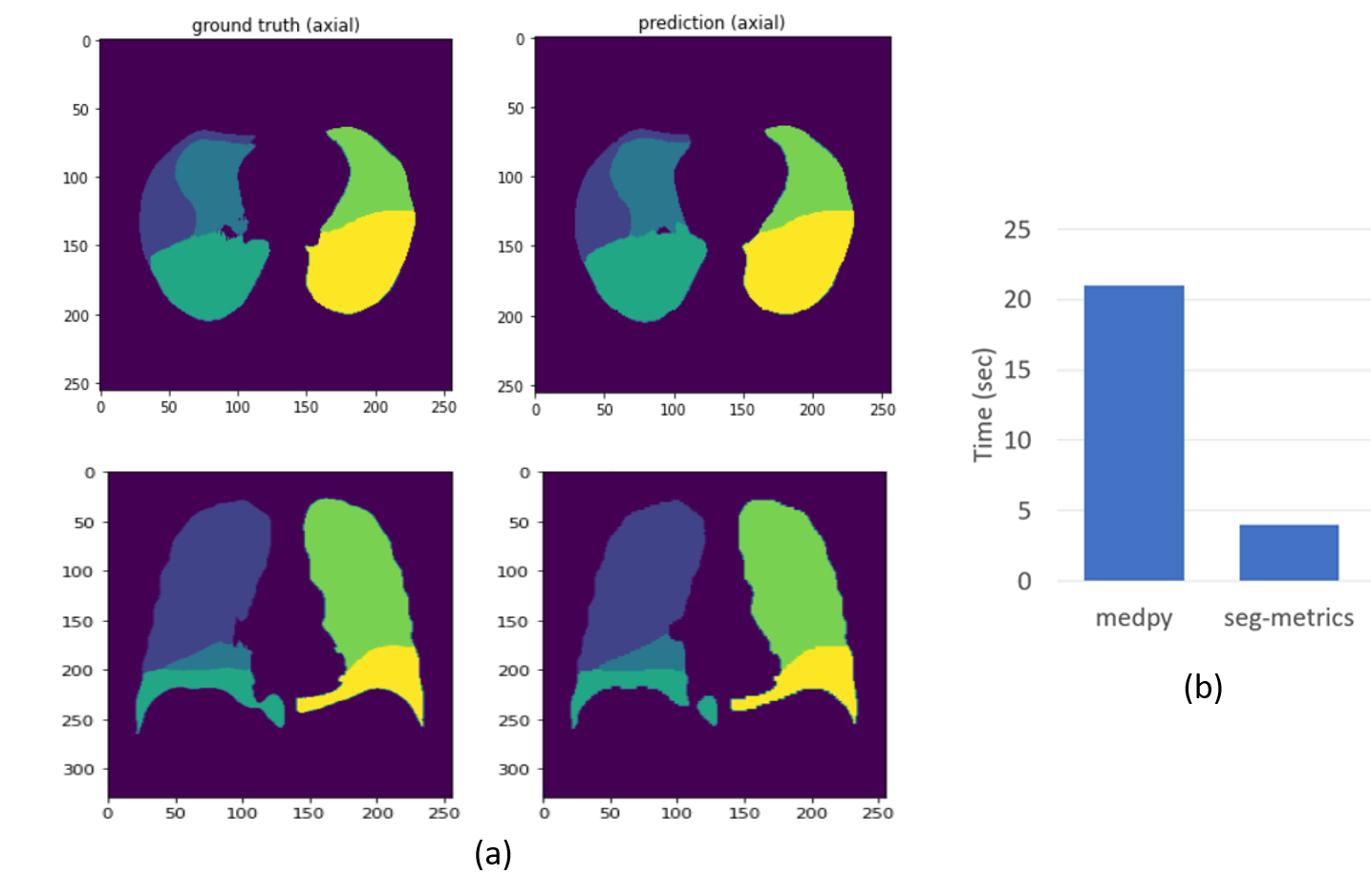}
        \caption{Performance comparison between \texttt{medpy} and \texttt{seg-metrics}. (a) Evaluated samples, a 3D lung lobe segmentation results (size: $256 \times 256 \times 256$). Left: ground truth (manually annotated lobes), right: prediction (automatically predicted lobes). (b) Time comparison for the calculation of 'hd', 'hd95' and 'msd'.}
        \label{fig:comp}
    \end{figure}
    
    \item \textbf{More convenient.} \texttt{seg-metrics} can calculate all different metrics in once in one function (shwon below)
    
    \begin{lstlisting}[frame=single]
gdth, pred = ...... # load two images
metrics = sg.write_metrics(labels=[1],
                  gdth_img=gdth,
                  pred_img=pred,
                  spacing=spacing,
                  metrics=['hd', 'hd95', 'msd']) # 3 outputs
\end{lstlisting}
    
    while \texttt{medpy} needs to call different functions multiple times which cost more code and time, because the calculation of each 'hd', 'hd95', and 'msd' will always recalculate the distance map which cost much time.
    
\begin{lstlisting}[frame=single]
hd = medpy.metric.binary.hd(result=pred, reference=gdth)
hd95 = medpy.metric.binary.hd95(result=pred, reference=gdth)
msd = medpy.metric.binary.asd(result=pred, reference=gdth)
\end{lstlisting}

    \item \textbf{More Powerful.} \texttt{seg-metrics} can calculate multi-label segmentation metrics and save results to .csv file in good manner, but \texttt{medpy} only provides \textbf{binary} segmentation metrics. For instance, if there are 5 labels for an image, our \texttt{seg-metrics} can calculate 5-label metrics by one-line command while \texttt{medpy} needs to at first convert 5-label image to five binary images, then calculate binary metrics one by one, 
\end{itemize}

\section{Limitation and future work}
Because of time limitation, there are still some space for the package to improve. 
\begin{itemize}
    \item Package name. The package name is "seg-metrics" currently, as the abbreviation of "segmentation metrics". But the dash sign "-" in the name introduced some confusion during the installing and usage of the package. Duing the installation, \texttt{pip install seg-metrics} is used. However, users need to used it by \texttt{import seg\_metrics}. The slight difference sometimes make new users confused and easy to make mistakes. This issue is because Python packaging system will automatically convert "\_" to "-" during the installing \cite{dash_underscore_email, dash_underscore_google, dash_underscore_overflow}). Because "segmetrics" has been used by other products, we may consider to change the package name to "metricseg", "metricsrater", "imagesegmetrics", etc. to avoid such issue in the future.
    
    \item Supported file type. Currently, the package supports most medical image formats with suffix of \texttt{.mhd, .mha, .nii, .nii.gz, .nrrd}, etc. Because we receive some users' requests, we will support more image formats (e.g. \texttt{.png, .jpg}) in the future.

    \item Usage guide. Currently, we just list the usage of different metrics, but we did not explain when to use which metrics. In the future, we hope to release a tutorial to users with some examples to show in which case, which metrics are preferable.
    
\end{itemize}

\section{Availability and requirements}
\begin{itemize}
    \item \textbf{Project name:} Seg-metrics
    \item \textbf{Project home page:} \url{https://github.com/Jingnan-Jia/segmentation_metrics}
    \item \textbf{Operating system(s):} Platform independent
    \item \textbf{Programming language:} Python
    \item \textbf{License:} MIT license
    \item \textbf{Any restrictions to use by non-academics:} none
\end{itemize}

\section*{Acknowledgments}
This author was supported by the China Scholarship Council No.202007720110 during the development of this package.

\printbibliography

\end{document}